\documentclass{IEEEtran}
\usepackage{cite}
\usepackage{amsmath,amssymb,amsfonts}
\usepackage{algorithmic}
\usepackage{graphicx,color}
\usepackage{textcomp}
\usepackage{academicons}  
\usepackage{hyperref}     
\usepackage{orcidlink}
\usepackage{graphicx}     
\usepackage{academicons}  
\usepackage{hyperref}     
\usepackage{xcolor}       
\usepackage[T1]{fontenc}
\usepackage{lmodern}
\usepackage{float}
\usepackage[skip=2pt]{caption} 
\usepackage{booktabs}   
\usepackage[linesnumbered,ruled,vlined]{algorithm2e}
\usepackage{orcidlink}
\usepackage{scalefnt}

\begin{document}

\title{Tackling Snow-Induced Challenges: Safe Autonomous Lane-Keeping with Robust Reinforcement Learning}

\author{
    Amin Jalal Aghdasian\textsuperscript{\scalebox{1.25}{\orcidlink{0009-0003-8482-1219}} 1},
    Farzaneh Abdollahi\textsuperscript{\scalebox{1.25}{\orcidlink{0000-0003-4957-987X}} 1},
    Ali Kamali Iglie\textsuperscript{2}\\
    \textsuperscript{1}Department of Electrical Engineering, Amirkabir University of Technology, Tehran, Iran\\
    \textsuperscript{2}Department of Mechanical Engineering, Amirkabir University of Technology, Tehran, Iran
}

\maketitle

\begin{abstract}
This paper proposes two new algorithms for the lane keeping system (LKS) in autonomous vehicles (AVs) operating under snowy road conditions. These algorithms use deep reinforcement learning (DRL) to handle uncertainties and slippage. They include Action-Robust Recurrent Deep Deterministic Policy Gradient (AR-RDPG) and end-to-end Action-Robust convolutional neural network Attention Deterministic Policy Gradient (AR-CADPG), two action-robust approaches for decision-making. In the AR-RDPG method, within the perception layer, camera images are first denoised using multi-scale neural networks. Then, the centerline coefficients are extracted by a pre-trained deep convolutional neural network (DCNN). These coefficients, concatenated with the driving characteristics, are used as input to the control layer. The AR-CADPG method presents an end-to-end approach in which a convolutional neural network (CNN) and an attention mechanism are integrated within a DRL framework. Both methods are first trained in the CARLA simulator and validated under various snowy scenarios. Real-world experiments on a Jetson Nano-based autonomous vehicle confirm the feasibility and stability of the learned policies. Among the two models, the AR-CADPG approach demonstrates superior path-tracking accuracy and robustness, highlighting the effectiveness of combining temporal memory, adversarial resilience, and attention mechanisms in AVs.
\end{abstract}


\section{INTRODUCTION}
\IEEEPARstart{A}{utonomous} vehicles can drive and perceive with minimum or no intervention at all from humans. They can increase safety, reduce accidents, and optimize traffic flow \cite{chen2022milestones}.
AVs rely on three fundamental operational components. The first component is perception, which involves recognizing and interpreting road signs and markings \cite{wang2023multi}. The second component is path planning, which focuses on determining a safe path from origin to destination. The third component is control, which adjusts steering, braking, and acceleration to ensure the vehicle moves smoothly and safely \cite{parekh2022review,abdollahian2024enhancing}.
Despite these advances, real-world use is challenged by uncertainties. At the perception layer, environmental factors such as adverse weather conditions can reduce the accuracy of the sensor \cite{zhang2023perception}. This can occur due to unreliable interpretations of the driving scene. Also, the control layer must face uncertainties that occur from dynamic road conditions, tire-road friction variability, and activation delays \cite{liu2023systematic}. Addressing these uncertainties is crucial to ensuring the robustness and safety of automated driving systems, especially in complex operating conditions.

\subsection{RELATED WORK}
The LKS is an autonomous system that uses sensor information to detect the environment and control a vehicle lateral motion. LKS aims to keep a vehicle in a lane and minimize orientation and lateral deviation with respect to the lane center. In the following, related works are presented for the perception and control layers of AVs.

\subsubsection{Some approaches in the Perception Layer}
Traditional line detection methods usually follow a sequential pipeline including image preprocessing, feature extraction, model fitting, and tracking. In the preprocessing stage, techniques such as colour correction in HSL space \cite{garg2022licent}, noise filtering including median, Gaussian, and FIR filters, thresholding \cite{amiriebrahimabadi2024comprehensive}, and colour space transformation \cite{tajjour2023novel} are used to enhance image clarity. Gradient-based methods like Canny edge detection and the Hough Transform are used for feature extraction \cite{akbari2021multilane}. Straight lines or more flexible geometric shapes, like parabolic curves \cite{paraschos2018using}, B-splines, and clothoid curves \cite{li2023trajectory}, are often used to model lane shapes. Although these classical methods perform well in structured environments, they are generally not robust to real-world challenges such as variable lighting conditions, sensor noise, and adverse weather conditions, all of which introduce uncertainty and reduce perception accuracy.

Recent works have significantly advanced the field of robust line segment detection through deep learning and hybrid models. In \cite{lu2021super}, Effective and Efficient Line Segment Detection was introduced. \cite{lin2024comprehensive} presented a comprehensive review of image line segment detection and description, proposing new taxonomies, benchmarking deep, hybrid, and conventional methods, and offering insights into their limitations and strengths across illumination, scale, and scene complexity. Complementing these efforts, LineDL was proposed in \cite{huang2023linedl} as a deep learning approach that processes images in a line-by-line fashion, aiming to reduce computational overhead while maintaining robust detection performance on mobile-scale hardware. Moreover, CNN architectures have significantly improved the robustness and efficiency of line segment detection, especially in cluttered or noisy environments. In \cite{teplyakov2022lsdnet}, a trainable CNN-based modification of the traditional Line Segment Detector, which integrates a line heatmap estimator and tangent field prediction module. The model validates line proposals by analysing geometric consistency and context-aware features, thereby reducing false positives and improving robustness in complex structural scenes.

\subsubsection{Some approaches in control layer}
There are various methods for the control layer in AVs. Generally, these methods are divided into model-based and model-free categories. 
In \cite{chen2019autonomous}, a linear quadratic regulator control method has been used for the lane-keeping system under sunny conditions. Despite the satisfactory performance of the system, any change in the front tire stiffness of the autonomous vehicle requires redesigning the controller. 
In \cite{dai2023tube}, a robust hybrid method using a Model Predictive Controller and a sliding mode controller is discussed to improve the accuracy and robustness of path tracking for an autonomous vehicle. This paper presents experimental results under three different road conditions, namely muddy, snowy, and icy roads. However, despite the good performance of classical controllers, they are often environment-dependent or require knowledge of the system dynamics \cite{lin2020comparison}. Therefore, their corresponding hyperparameters should be tuned appropriately for each environment to achieve the desired behaviour, or the system dynamics should be identified.

Various studies of DRL algorithms have explored the application of these algorithms in AVs control systems \cite{huang2022survey}. In \cite{perez2022deep}, a Deep Q-Learning algorithm was applied for vehicle control in sunny conditions. However, due to the algorithm limitation in handling continuous action spaces, the results were suboptimal. To address this issue, the Deep Deterministic Policy Gradient (DDPG) algorithm was introduced, showing improved performance. Addressing the robustness of policies, the concept of adversarial attacks was explored. In \cite{ilahi2021challenges}, two adversarial attack models are introduced: the Myopic Action Space Attack and the Look-Ahead Action Space Attack, with the latter demonstrating a more significant impact on degrading DRL performance.
In \cite{klima2019robust}, a robust policy framework against adversarial attacks has been proposed using an enhanced Temporal Difference learning method. This approach draws parallels to learning that maximizes the worst-case return in two-player zero-sum games. It focuses instead on minimizing the opponent’s action space to strengthen control performance.
In \cite{tessler2019action}, a robust hierarchical RL framework was introduced, utilizing a fixed set of low-level skills to enable efficient learning in complex environments. The proposed approach separates skill learning from task learning, allowing the agent to focus on high-level decision-making while reusing pre-trained primitives. In \cite{deshpande2021robust}, a robust DRL approach was proposed using the Action-Robust Deep Deterministic Policy Gradient (AR-DDPG) algorithm. This method integrates the Robust Markov Decision Process (RMDP) framework with adversarial training, where an adversary perturbs the control actions with a tunable probability to simulate environmental and parametric uncertainties.  

These studies highlight the evolution of DRL-based control for AVs, emphasizing the role of robustness and adversarial resilience in improving the stability and adaptability of control strategies.

\subsection{MOTIVATIONS Of THE PAPER}
AV in snowy weather presents a significant challenge for self-driving systems, because it introduces considerable uncertainties in both vision and control systems. These uncertainties can have a significant impact on the safety and accuracy of autonomous systems. Many existing methods are not fully robust for real-world scenarios, especially in complex environmental conditions like snow. Thus, there is an essential need to develop algorithms that can maintain robust performance under such uncertain conditions. 

\subsection{CONTRIBUTIONS OF THE PAPER}
The main contributions of this paper are as follows:\\
1. A reward function is proposed to improve the LK behavior under snow-induced uncertainty. This function penalizes lateral deviation, excessive control actions, and unsafe path tracking, ensuring smoother and safer control responses in low-friction conditions.\\
2- Two novel robust DRL algorithms, AR-RDPG and AR-CADPG, are introduced. AR-RDPG incorporates temporal memory via recurrent neural networks to handle partial observability. Moreover, AR-CADPG integrates spatial attention mechanisms and adversarial robustness in an end-to-end visual control pipeline.\\
The proposed methods are trained and validated in the CARLA simulator under various snow scenarios and tested on a Jetson Nano-based robotic car. The results demonstrate that AR-CADPG achieves the best performance in terms of tracking accuracy, robustness, and stability, in simulation as well as on real hardware.\\

The remainder of this paper is organized as follows: Section II defines the problem formulation for lane-keeping in snowy environments using a robust RL framework. Section III presents the proposed methodology, including the AR-RDPG and AR-CADPG algorithms and the perception pipeline. Section IV provides simulation results and comparative evaluations of the proposed controllers. Section V details the real-world implementation and experimental validation using a Jetson Nano-based autonomous vehicle platform. Finally, Section VI concludes the paper.

\section{Problem Definition}
AV in snowy environments introduces significant challenges due to the presence of perceptual and control uncertainties. Snowfall and road coverage, unclear lane markings, and under-the-influence sensor inputs complicate visual perception. Simultaneously, variable friction, tire slippage, and unpredictable external disturbances introduce uncertainty into the vehicle control. These challenges underscore the need for control strategies that are robust to perceptual uncertainty as well as action execution uncertainty.

To address the above challenge, the LK task is formulated as a robust RL problem within the framework of a RMDP. The RMDP extends the conventional Markov Decision Process by incorporating uncertainty into the environment dynamics through bounded perturbations, thereby modeling adversarial uncertainty that may degrade policy performance. For that purpose, the RMDP is defined by the tuple $\mathcal{M} = (S, A, P, R, \gamma)$, where:
\begin{itemize}
	\item $S$ denotes a continuous state space comprising features 
	\item $A$ represents a continuous action space corresponding to control inputs
	\item $P$ is a bounded transition model that captures environmental uncertainty
	\item $R$ is the reward function
	\item $\gamma \in (0, 1)$ is the discount factor
\end{itemize}

In this work, robustness is modeled in the action space, following the Action-Robust Markov Decision Process (AR-MDP) framework.

The objective is to learn a robust policy $\pi^*$ that maximizes the expected return against the worst-case adversarial policy:

\begin{equation}
	\pi^* = \arg \max_{\pi} \min_{\bar{\pi}} \mathbb{E}_{\pi_{\text{mix}}} \left[ \sum_{t=0}^{\infty} \gamma^t r(s_t, a_t) \right]
\end{equation}

\noindent
where:
\begin{itemize}
	\item $\pi^*$ denotes the optimal policy of the agent, which maximizes the expected return
	\item $\bar{\pi}$ represents the adversarial policy, which introduces uncertainty into the agent action execution
	\item $\pi_{\text{mix}}$ is the mixed policy resulting from the combination of $\pi$ and $\bar{\pi}$
\end{itemize}

\section{Methology}

\subsection{Action-Robust Recurrent Deep Deterministic Policy Gradient}

\paragraph{Perception Layer}

The perception layer is designed for lane detection and the extraction of centerline coefficients that define the midline between two road lanes. A front camera sensor is employed to capture road imagery for this purpose. However, in the context of snowy environments, lane detection becomes particularly challenging.

\begin{figure}[htbp]
	\centering
	\includegraphics[width=\linewidth]{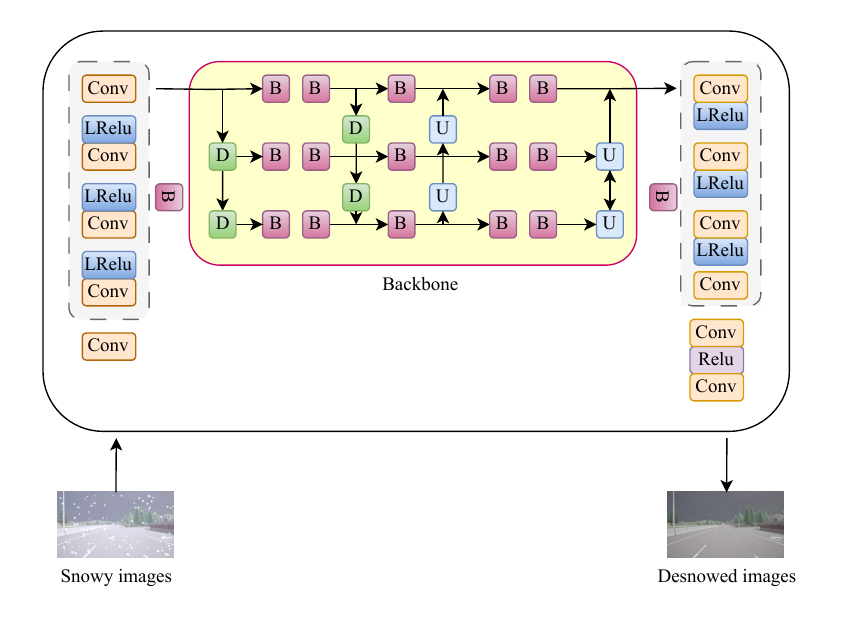}
	\caption{Deep Multi-Scale Dense Network for image desnowing. Here, D and U denote down-sampling and up-sampling modules, respectively~\cite{zhang2021deep}.}
	\label{Fig1}
\end{figure}

To solve this problem, the raw images obtained from the front camera are first processed through a snow removal pipeline. A deep multi-scale dense network, as proposed in ~\cite{zhang2021deep} and shown in Fig.~\ref{Fig1}, is used to remove snowflakes from the images. This network enhances desnowing performance, and a stage multi-scale refinement module, DDMSNet, is applied. This network incorporates semantic and geometric cues through attention mechanisms, resulting in a final, high-quality snow-free image suitable for tasks.

\begin{figure}[htbp]
	\centering
	\includegraphics[width=\linewidth]{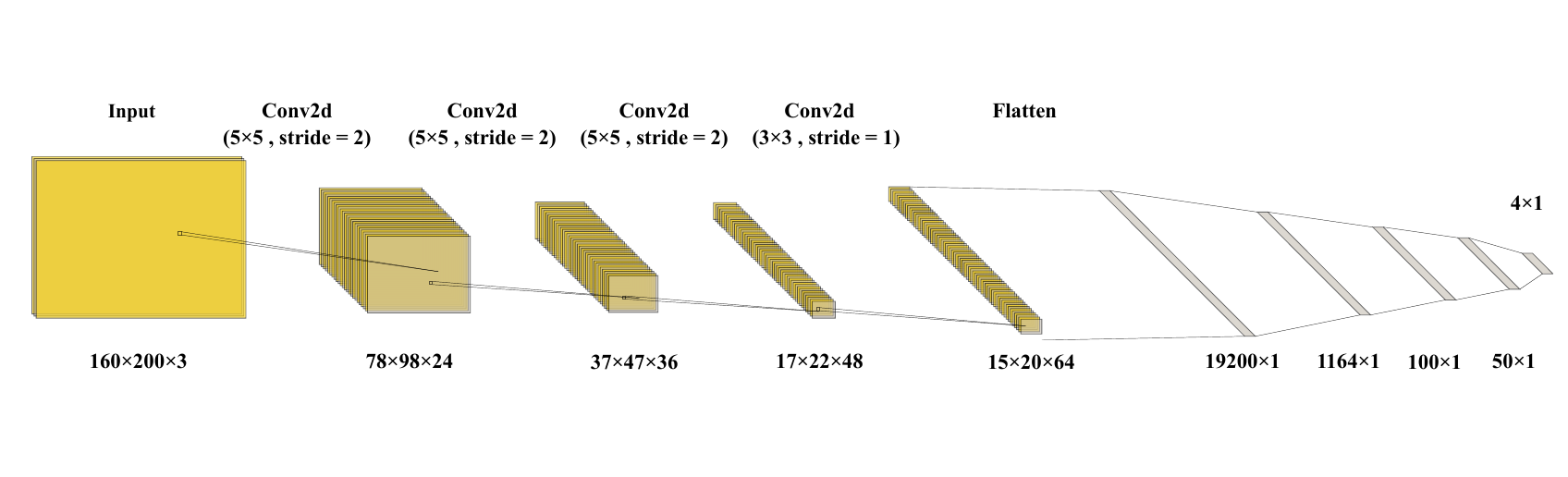}
	\caption{Architecture of the DCNN for estimating centerline coefficients~.}
	\label{Fig2}
\end{figure}

After snow removal, the cleared road images are fed to a DCNN, whose architecture is shown in Fig.~\ref{Fig2}. The main goal of this network is to estimate the coefficients that define the geometric shape of the lane centerline. These coefficients are the parameters of a polynomial curve ($c_0, c_1, c_2, c_3$) that show where the road is in relation to the vehicle current position, how it curves, and which way it goes.
The perception pipeline makes the snow-cleared image that goes into the DCNN. The network is trained ahead of time to deal with different visual conditions, and it can generalize well even when left or right lane markings are common in snowy areas. The network uses a series of convolution and fusion layers to get spatial and contextual features from the road scene. Then, fully connected layers regress the coefficients of the final line model.
These predicted coefficients are then added to the agent state representation for the controller that uses RL. They give important information for the LK task, like how curved the road segment is ahead and where the vehicle is in relation to the ideal centerline trajectory. This enhanced state lets the control policy make better and more stable driving choices, especially when visibility is low or unclear.

\paragraph{RNN-Based Action-Robust Policy Formulation}

To enable robustness in partially observable driving scenarios, this work extends the Action-Robust Deep Deterministic Policy Gradient (AR-DDPG) framework~\cite{tessler2019action} by incorporating recurrent neural networks (RNNs) into the actor, critic, and adversarial branches. These recurrent architectures preserve a hidden state vector that captures temporal dependencies, enabling decision-making based on sequences of observations.

\begin{figure}[htbp]
	\centering
	\includegraphics[width=0.8\linewidth]{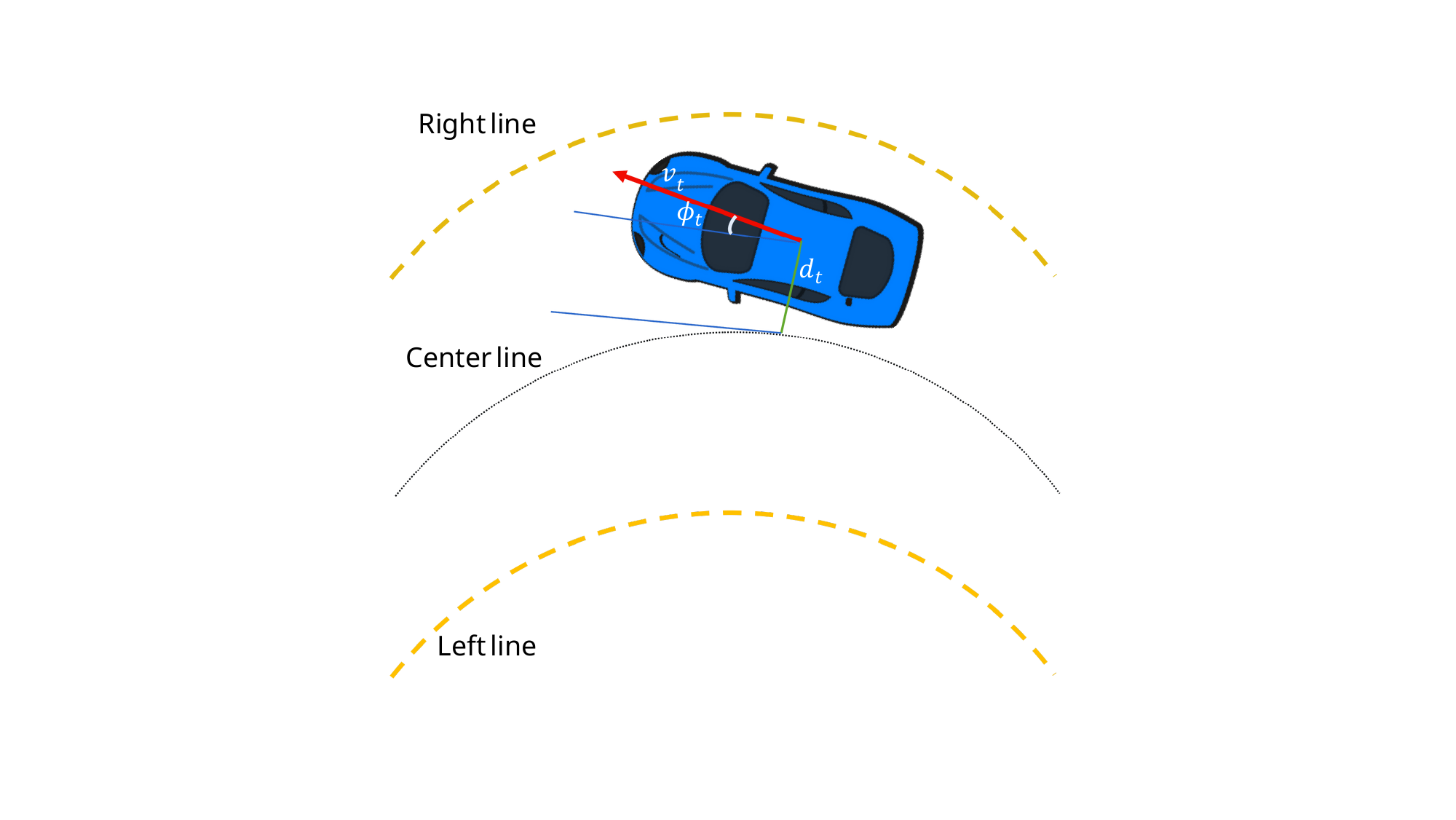}
	\caption{Kinematic quantities used in the observation vector: lateral deviation $d_t$, heading angle $\phi_t$, and forward velocity $v_t$.}
	\label{fig_lat_err}
\end{figure}

At each time step $t$, the input observation $o_t$ is constructed from key kinematic variables: the vehicle lateral deviation $d_t$, heading angle $\phi_t$, and forward velocity $v_t$. These variables form the observation vector $o_t = [d_t, \phi_t, v_t]$ and serve as inputs to the RNN-based networks. The geometric meaning of these features is visualized in Fig.~\ref{fig_lat_err}.

The hidden state $h_t$ is updated via a simple recurrent unit:
\begin{equation}
	h_t = \tanh(W_h h_{t-1} + W_o o_t + b)
\end{equation}
where $W_h$ and $W_o$ denote the recurrent and input weights, respectively, and $b$ is a bias vector. This hidden state $h_t$ serves as a compact encoding of the temporal context and is shared among the actor, critic, and adversarial networks.

To account for uncertainty in the control channel, the actual action executed by the system is a combination of the agent action $\mu(h_t)$ and an adversarial action $\bar{\mu}(h_t)$:
\begin{equation}
	\tilde{a}_t = (1 - \alpha)\mu(h_t) + \alpha \bar{\mu}(h_t)
\end{equation}
where $\alpha \in [0,1]$ adjusts the influence of the adversary.

The critic estimates the action-value function based on the hidden state and the perturbed action:
\begin{equation}
	Q(h_t, \tilde{a}_t) \approx \mathbb{E} \left[ \sum_{k=0}^{\infty} \gamma^k r_{t+k} \right]
\end{equation}
The critic is trained using the target value:
\begin{equation}
	y_t = r_t + \gamma Q_{\theta'}(h_{t+1}, \mu_{\phi'}(h_{t+1}))
\end{equation}
where $\theta'$ and $\phi'$ are the parameters of the target critic and actor networks.

The actor is updated using the deterministic policy gradient:
\begin{equation}
	\nabla_{\phi} J(\phi) = \mathbb{E}_{h_t} \left[ \nabla_a Q(h_t, a) \big|_{a = \mu(h_t)} \cdot \nabla_{\phi} \mu(h_t) \right]
\end{equation}
While the adversarial policy is trained to degrade performance:
\begin{equation}
	\nabla_{\bar{\phi}} J(\bar{\phi}) = -\mathbb{E}_{h_t} \left[ \nabla_a Q(h_t, a) \big|_{a = \bar{\mu}(h_t)} \cdot \nabla_{\bar{\phi}} \bar{\mu}(h_t) \right]
\end{equation}

\begin{figure}[htbp]
	\centering
	\includegraphics[width=\linewidth]{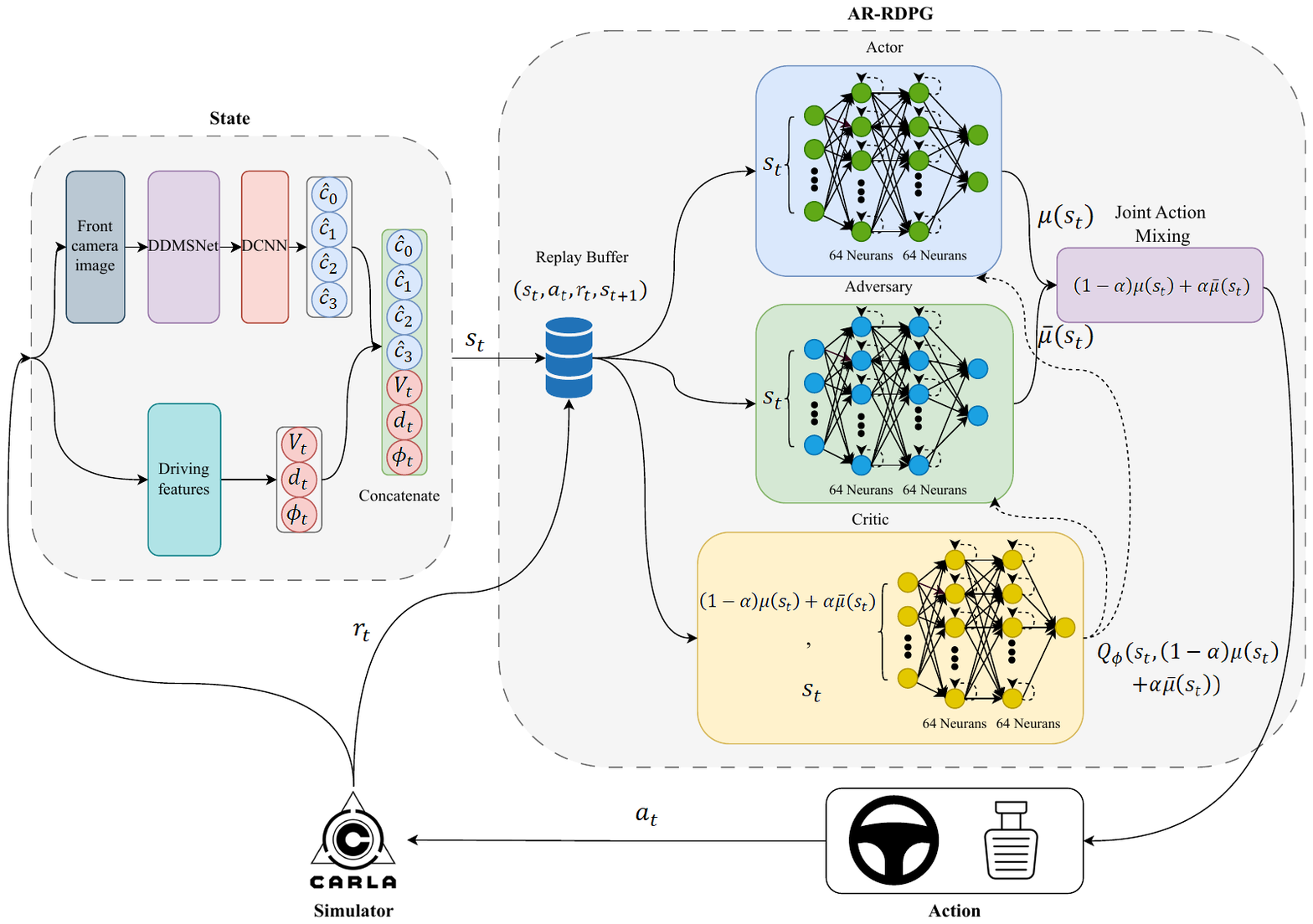}
	\caption{The proposed AR-RDPG framework combining deep perception, recurrent policy learning, and robust action execution via agent–adversary interaction.}
	\label{fig_ar_rdp}
\end{figure}

Fig.~\ref{fig_ar_rdp} presents the overall AR-RDPG pipeline, integrating perception, recurrent policy learning, and robust action formulation. The architecture begins by processing camera and sensor data, followed by feature extraction and control generation using RNNs.

The corresponding training procedure is outlined in Algorithm~\ref{alg_ar_rdp}, where RNN-based actor and critic networks interact under adversarial influence and are updated using temporal mini batches from the replay buffer.

\begin{algorithm}[h]
	\caption{RNN-Based Action-Robust Deep Deterministic Policy Gradient (AR-RDPG)}
	\label{alg_ar_rdp}
	\KwIn{Replay buffer $\mathcal{D}$, actor $\mu(h_t)$, adversary $\bar{\mu}(h_t)$, critic $Q(h_t, a)$, learning rates $\alpha_{\mu}, \alpha_{\bar{\mu}}, \alpha_Q$, mixing factor $\alpha$}
	\KwOut{Optimized policies $\mu^*, \bar{\mu}^*$}
	
	Initialize parameters $\theta_Q$, $\phi_\mu$, $\bar{\phi}$ and target networks $\theta_Q'$, $\phi_\mu'$\;
	Initialize RNN hidden states $h_0$ for all networks\;
	Initialize replay buffer $\mathcal{D}$\;
	
	\For{episode = 1 \KwTo M}{
		Initialize observation $o_0$ and hidden state $h_0$\;
		\For{t = 0 \KwTo T}{
			Update hidden state: $h_t = \tanh(W_h h_{t-1} + W_o o_t + b)$\;
			Generate agent action: $a_t^\mu = \mu(h_t) + \mathcal{N}_t$\;
			Generate adversarial action: $a_t^{\bar{\mu}} = \bar{\mu}(h_t)$\;
			Mix actions: $\tilde{a}_t = (1 - \alpha)a_t^\mu + \alpha a_t^{\bar{\mu}}$\;
			Execute $\tilde{a}_t$, observe $r_t$, $o_{t+1}$\;
			Store transition $(o_t, h_t, \tilde{a}_t, r_t, o_{t+1})$ in $\mathcal{D}$\;
		}
		Sample minibatch from $\mathcal{D}$\;
		\ForEach{sample $(o, h, \tilde{a}, r, o')$}{
			Update $h'$ from $o'$ using the RNN\;
			Compute target: $y = r + \gamma Q_{\theta_Q'}(h', \mu_{\phi'}(h'))$\;
			Update critic: $\theta_Q \leftarrow \theta_Q - \alpha_Q \nabla_{\theta_Q}(Q(h, \tilde{a}) - y)^2$\;
			Update actor: $\phi_\mu \leftarrow \phi_\mu + \alpha_\mu \nabla_{\phi_\mu} Q(h, a)\big|_{a = \mu(h)}$\;
			Update adversary: $\bar{\phi} \leftarrow \bar{\phi} - \alpha_{\bar{\mu}} \nabla_{\bar{\phi}} Q(h, a)\big|_{a = \bar{\mu}(h)}$\;
		}
		Soft update targets:\;
		\hspace{1em} $\theta_Q' \leftarrow \tau \theta_Q + (1 - \tau)\theta_Q'$\;
		\hspace{1em} $\phi' \leftarrow \tau \phi_\mu + (1 - \tau)\phi'$\;
	}
\end{algorithm}

\subsection{End-to-End Action-Robust CNN-Attention DDPG}

To extend robustness to visual perception under adverse conditions, we propose an end-to-end architecture, AR-CADPG. This model jointly learns to extract visual features, refine them using attention, and predict control actions directly from raw images and vehicle kinematics.

\paragraph{Observation Encoding and Feature Fusion}

The agent receives a camera image \( I_t \in \mathbb{R}^{H \times W \times 3} \), in addition to the standard kinematic state variables: forward velocity \( v_t \), lateral deviation \( d_t \), and heading angle \( \phi_t \). In contrast to the earlier models, visual input is processed end-to-end.

A CNN encodes the image into a spatial feature map \( F \in \mathbb{R}^{C \times H' \times W'} \). To suppress irrelevant regions, a spatial attention module generates an attention mask \( M_s \in [0,1]^{H' \times W'} \), computed as:
\begin{equation}
	M_s = \sigma\left(f^{7 \times 7} \left( \text{Concat}[\text{AvgPool}(F), \text{MaxPool}(F)] \right)\right)
\end{equation}
where \( f^{7 \times 7} \) is a convolution with a \( 7 \times 7 \) kernel and \( \sigma(\cdot) \) is the sigmoid function.

The attended feature map is obtained by element-wise multiplication:
\begin{equation}
	F' = F \odot M_s
\end{equation}

Flattened features from \( F' \) are passed through a fully connected layer to obtain the visual embedding \( z_v \). The kinematic input is separately projected to an embedding \( z_k = \text{FC}([v_t, d_t, \phi_t]) \). These embeddings are fused as:
\begin{equation}
	z = \text{FC}(\text{Concat}(z_v, z_k))
\end{equation}

\paragraph{Policy Learning and Robust Control}

\begin{figure}[htbp]
	\centering
	\includegraphics[width=\linewidth]{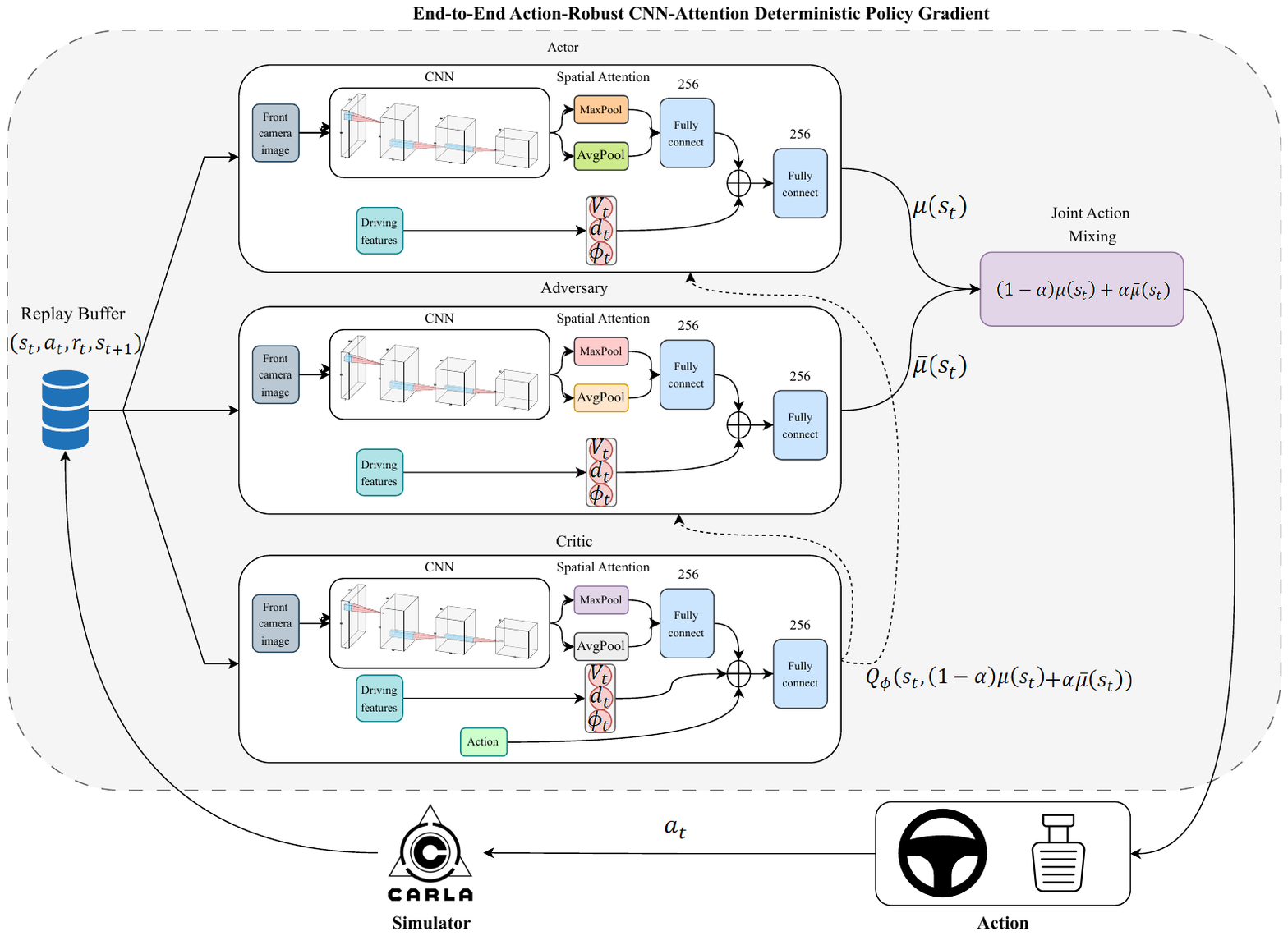}
	\caption{The proposed AR-CADPG architecture integrating visual processing, spatial attention, feature fusion, and robust policy generation.}
	\label{fig_end2end_arch}
\end{figure}

The overall architecture is shown in Fig.~\ref{fig_end2end_arch}. The fused feature vector \( z \) is passed through the actor network to produce a deterministic action \( \mu_t = \mu(s_t) \in \mathbb{R}^2 \). Robustness to execution uncertainty is achieved by introducing an adversarial policy and applying a mixing strategy at the action level. The critic network estimates the action-value function \( Q(s_t, \tilde{a}_t) \) for the mixed action.

\paragraph{Training Procedure}

The AR-CADPG framework is trained using off-policy temporal difference learning, similar to AR-RDPG, with raw image observations processed end-to-end. The actor, critic, and adversary networks are updated via deterministic policy gradients. The full training algorithm is stated in Algorithm~\ref{alg_ar_cnnattdpg}.

\begin{algorithm}[h]
	\caption{End-to-End Action-Robust CNN-Attention DDPG (AR-CADPG)}
	\label{alg_ar_cnnattdpg}
	\KwIn{Replay buffer \( \mathcal{D} \), actor \( \mu(s_t) \), adversary \( \bar{\mu}(s_t) \), critic \( Q(s_t, a) \), learning rates \( \alpha_{\mu}, \alpha_{\bar{\mu}}, \alpha_Q \), mixing factor \( \alpha \)}
	\KwOut{Trained actor and adversarial policies \( \mu^*, \bar{\mu}^* \)}
	
	Initialize parameters \( \theta_Q, \phi_\mu, \bar{\phi} \) and target networks \( \theta_Q', \phi_\mu' \)\;
	Initialize replay buffer \( \mathcal{D} \)\;
	
	\For{episode = 1 \KwTo M}{
		Initialize state \( s_0 = (I_0, v_0, d_0, \phi_0) \)\;
		\For{t = 0 \KwTo T}{
			Extract visual features \( F \) from \( I_t \) via CNN\;
			Apply spatial (optionally channel) attention to obtain \( F' \)\;
			Fuse \( F' \) with kinematics to get input \( z \)\;
			Compute agent action \( a_t^{\mu} = \mu(s_t) + \mathcal{N}_t \)\;
			Compute adversary action \( a_t^{\bar{\mu}} = \bar{\mu}(s_t) \)\;
			Mix actions: \( \tilde{a}_t = (1 - \alpha)a_t^{\mu} + \alpha a_t^{\bar{\mu}} \)\;
			Execute \( \tilde{a}_t \), observe reward \( r_t \), next state \( s_{t+1} \)\;
			Store transition \( (s_t, \tilde{a}_t, r_t, s_{t+1}) \) in \( \mathcal{D} \)\;
		}
		
		Sample mini-batch from \( \mathcal{D} \)\;
		\ForEach{sample \( (s, \tilde{a}, r, s') \)}{
			Extract features from \( s \) and \( s' \)\;
			Compute target: \( y = r + \gamma Q_{\theta_Q'}(s', \mu_{\phi_\mu'}(s')) \)\;
			Update critic: \( \theta_Q \leftarrow \theta_Q - \alpha_Q \nabla_{\theta_Q}(Q(s, \tilde{a}) - y)^2 \)\;
			Update actor: \( \phi_\mu \leftarrow \phi_\mu + \alpha_\mu \nabla_{\phi_\mu} Q(s, a)\big|_{a = \mu(s)} \)\;
			Update adversary: \( \bar{\phi} \leftarrow \bar{\phi} - \alpha_{\bar{\mu}} \nabla_{\bar{\phi}} Q(s, a)\big|_{a = \bar{\mu}(s)} \)\;
		}
		
		Soft update targets:\;
		\hspace{1em} \( \theta_Q' \leftarrow \tau \theta_Q + (1 - \tau)\theta_Q' \)\;
		\hspace{1em} \( \phi_\mu' \leftarrow \tau \phi_\mu + (1 - \tau)\phi_\mu' \)\;
	}
\end{algorithm}

\section{Simulation Results} \label{sim_res}
\begin{figure}[htbp]
\centering
\includegraphics[width=0.45\linewidth]{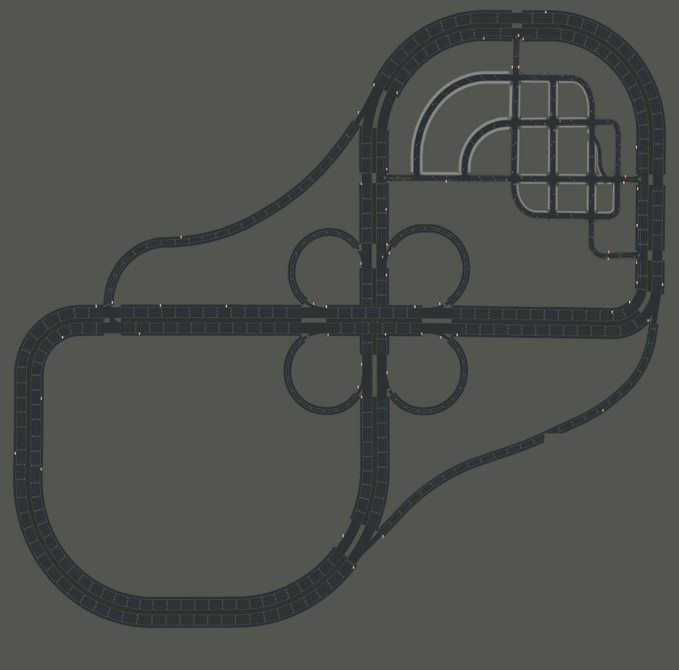}
\includegraphics[width=0.45\linewidth]{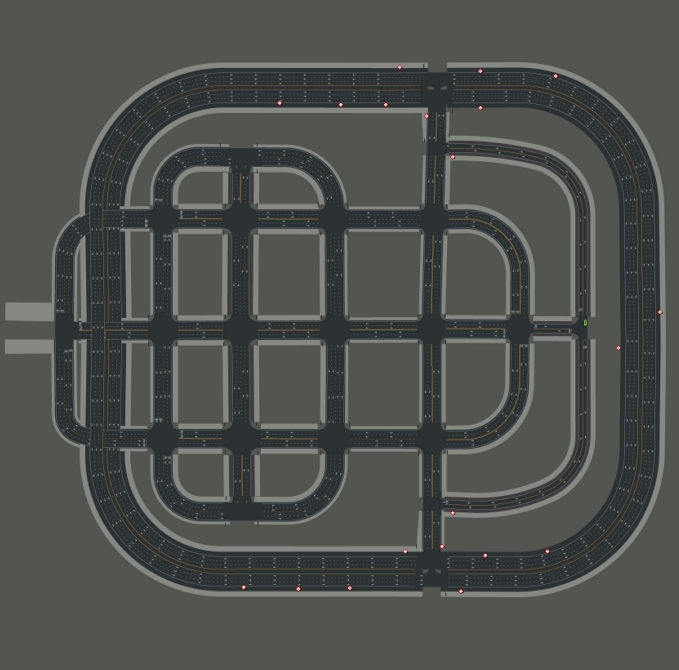} \\[1ex]
\includegraphics[width=0.45\linewidth]{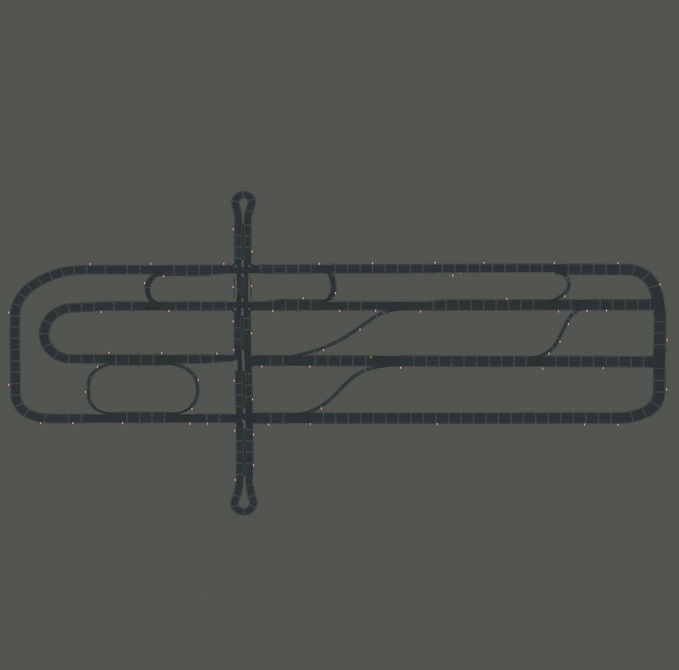}
\includegraphics[width=0.45\linewidth]{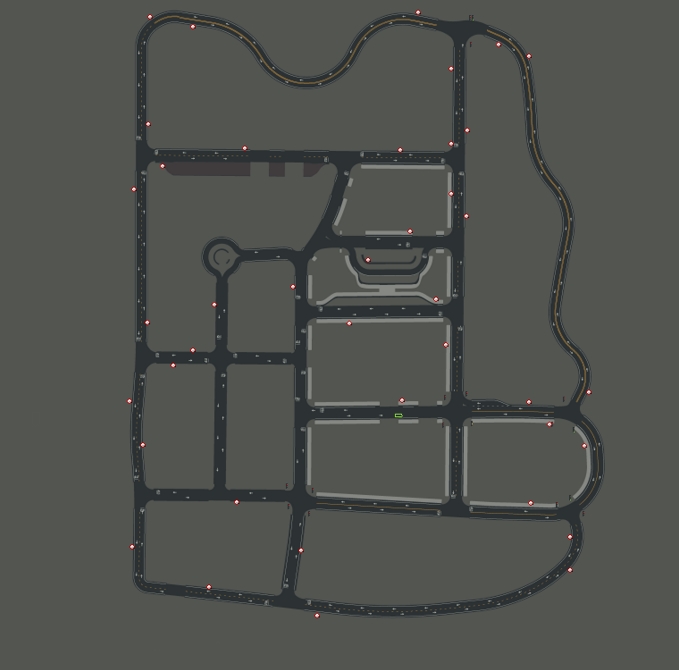}
\caption{Bird’s-eye view of the CARLA towns and their drivable routes used for training and validation. The four selected maps provide diverse road structures and driving scenarios to evaluate the generalization capability of the proposed models.}
\label{fig:town_maps}
\end{figure}

The proposed DRL algorithms are first trained until they reach convergence through episodic interactions in the CARLA simulator. CARLA is an open-source platform designed specifically for creating, training, and testing self-driving systems in highly realistic and easily controlled environments. Monte Carlo methods are then used to test each learned policy independently, providing statistically sound estimates of their effectiveness. In RL research, the best practice is to separate the learning and evaluation phases into two steps: training and policy verification. Moreover, both the training and validation processes use the map scenarios shown in Fig.~\ref{fig:town_maps}. These scenarios provide a common spatial environment for both processes.

During validation, the simulator runs $M$ episodes, each consisting of $T$ steps. To ensure that every route is different, an A*-based global planner selects random start and goal points at the beginning of each episode. The sensors are continuously monitored for crashes and lane changes. An episode terminates, and the car is reset to a valid starting position if either event is triggered. Otherwise, the episode proceeds as planned. In this way, each agent is tested in a range of driving situations, making the evaluation more reliable and applicable to other scenarios.

All tests are conducted on a desktop workstation equipped with a 12\textsuperscript{th} Gen Intel Core i9-12900HX CPU and an NVIDIA GeForce RTX 3080 Ti GPU with 16 GB of VRAM. This rigorous validation strategy enhances confidence in the generalizability and robustness of the learned policies, particularly in novel and challenging scenarios.

\vspace{1\baselineskip}

\begin{itemize}
    \item \textbf{Training stage} 
\end{itemize}

The first three algorithms use a trained DCNN as their perception module. The pre-training dataset contains 1000 images of snow and 1000 images of sunny conditions. Initially, the traditional Clothoid-based method labels the images by extracting the left and right lane boundaries from front-facing camera views, ensuring that the lanes are fully visible. This method performs well even on curved roads. By averaging the lane boundaries, the centerline is obtained.
A third-degree polynomial is then fitted to the extracted centerline coordinates using the rules of Clothoid curve fitting, which are commonly employed in road geometry to ensure smooth transitions. The four polynomial coefficients ($c_0$ to $c_3$) serve as training labels for the DCNN. The goal of training is to teach the model to predict the coefficients ($\hat{c}_0$ to $\hat{c}_3$) when lane markings are only partially or entirely hidden. The predicted lane parameters are then used as input states for the DRL algorithms.

To evaluate performance under adverse weather conditions, four DRL controllers—DDPG \cite{perez2022deep}, AR-DDPG \cite{tessler2019action}, AR-RDPG, and AR-CADPG—are trained. To reduce noise in the recorded data and highlight underlying learning trends, a moving average filter is applied to the performance metrics over time. The training process continues until the policies of every model synchronize, ensuring stable and safe driving behavior in snowy and volatile conditions. The finalized policies are then used to assess the results.
A tire–road friction coefficient of 0.6 is used during training to simulate conditions similar to snowy and slippery roads \cite{ojala2024road}. A classic image processing technique is applied to add artificial snowflakes to the scene, making it more challenging for the agent to perceive its surroundings. This controlled and realistic modification further reduces the clarity of the agent’s field of view. Additionally, lane perception uncertainty is introduced by randomly removing one or two lane markers on one or both sides of a lane during training. This strategy simulates situations in which snow partially or completely covers road markings, making the learned policy more robust under occlusion and reduced visibility. Table~\ref{tab:training_hyperparams} summarizes the hyperparameter values used during training.

\begin{table}[H]
	\centering
	\caption{Training Hyperparameters}
	\label{tab:training_hyperparams}
	\begin{tabular}{@{}ll@{}}
		\toprule
		\textbf{Parameter} & \textbf{Value} \\
		\midrule
		Tire–road friction coefficient        & $0.6$ \\
		Discount factor ($\gamma$)           & $0.95$ \\
		Antagonistic action probability ($\alpha$) & $0.1$ \\
		Learning rate $\eta_{(\pi, \bar{\pi})}$ & $2 \times 10^{-5}$ \\
		Learning rate $\eta_Q$               & $2 \times 10^{-4}$ \\
		Smoothness penalty weight ($\lambda_1$) & $0.1$ \\
		Throttle penalty weight ($\lambda_2$)    & $0.05$ \\
		Neural network optimizer             & Adam \\
		Replay buffer size $|\mathcal{R}|$   & $8 \times 10^{5}$ \\
		\bottomrule
	\end{tabular}
\end{table}

\begin{figure}[htbp]
	\centering
	\includegraphics[width=\linewidth]{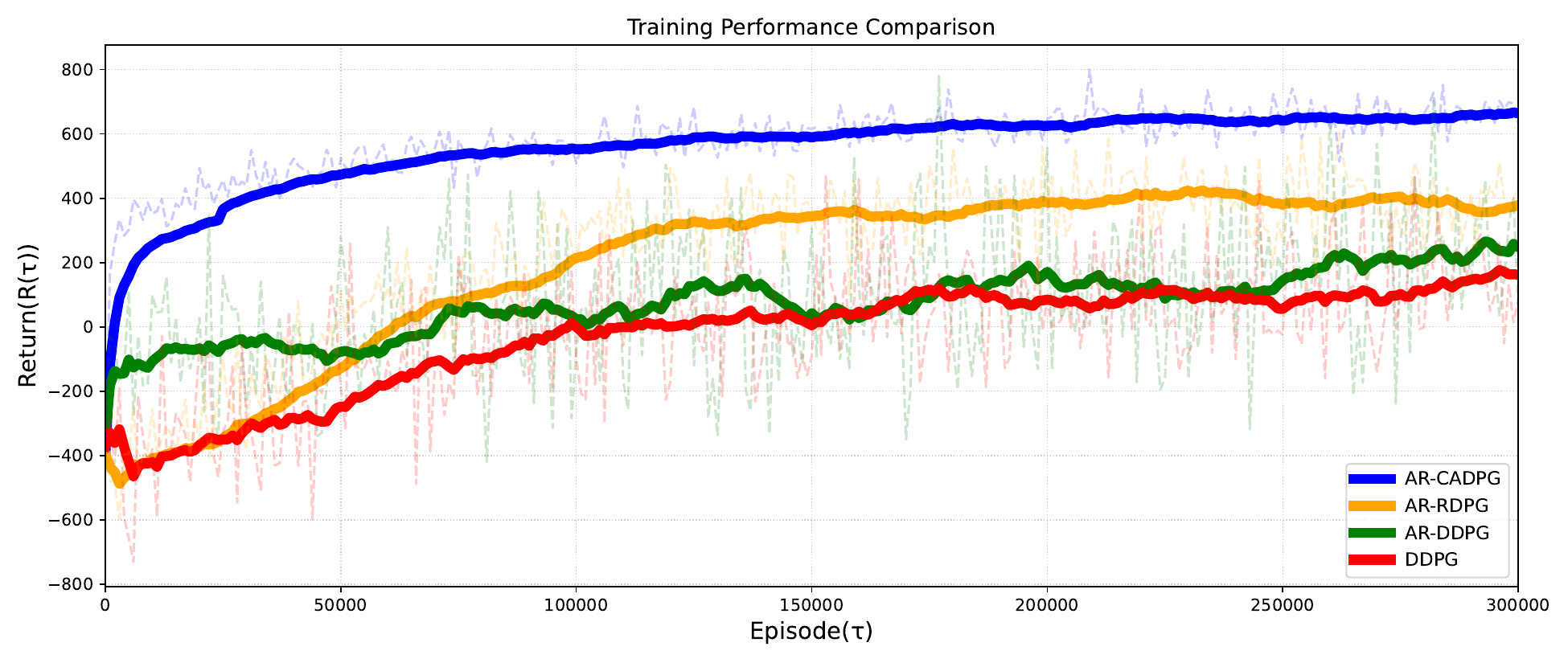}
	\caption{Training performance comparison of four DRL algorithms.}
	\label{fig:training}
\end{figure}

Fig.~\ref{fig:training} shows how the episodic return changes over time for each of the four models. A moving average filter smooths the performance curves, eliminating high-frequency noise and highlighting the underlying learning trends. All models demonstrate convergent behavior, as indicated by the stabilization of the return values when training is complete. Among the models, AR-CADPG achieves the highest final return, indicating faster learning and superior performance under adversarial conditions. Its awareness of the environment and resilience to attacks further enhance its potential for strong generalization.
AR-RDPG also performs well, with smooth learning dynamics and timely reward acquisition. The incorporation of recurrent layers enhances the model’s capacity to handle temporal dependencies more effectively in partially observable or noisy settings. In contrast, DDPG exhibits lower final returns and a reward curve with high variability, particularly during the initial training epochs, indicating a more volatile learning process and greater susceptibility to environmental uncertainties. AR-DDPG improves upon DDPG but remains inferior to both the recurrent and context-aware versions. These findings support the conclusion that incorporating adversarial robustness, recurrence, and contextual information in policy design improves learning stability and policy performance under snowfall and limited visibility.

\begin{itemize}
    \item \textbf{Validation Procedure and Metrics}
\end{itemize}

\begin{figure}[htbp]
  \centering
  \includegraphics[width=\linewidth]{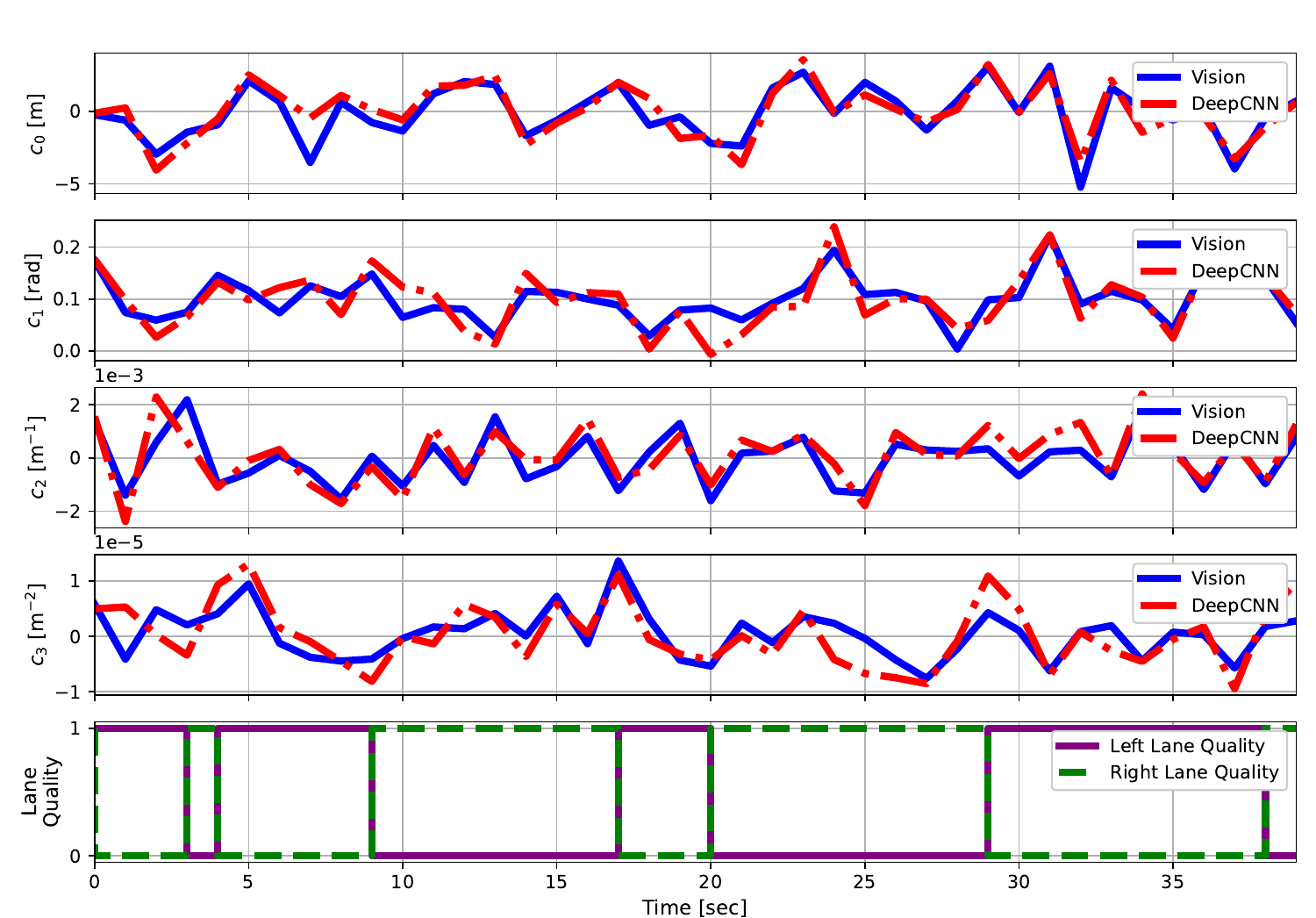}
  \caption{Validation results showing accurate centerline tracking by the trained DCNN, even in cases with missing or occluded lane markings.}
  \label{fig:validation}
\end{figure}

Validation tests demonstrate that the DCNN maintains reliable lane tracking performance, even in scenarios where lane markings are absent, such as snow-covered roads. As shown in Fig.~\ref{fig:validation}, the network effectively infers the centerline using learned visual cues, confirming the robustness of the Clothoid-guided polynomial labeling approach.

To thoroughly evaluate how well the proposed RL controllers maintain lane keeping, 50 different and randomly generated routes are used in the CARLA simulator. A global planner based on A* selects each route to ensure a variety of road shapes, including straight sections, curves, and intersections, under snowy and low-visibility conditions.

To further assess algorithm robustness, the tire–road friction coefficient is set to 0.5 during validation. This configuration simulates low-traction surfaces such as icy or compacted snow-covered roads, introducing additional challenges for lateral vehicle control.

Regression-based metrics evaluate how closely the ego vehicle follows the lane centerline in each episode. Specifically, the Root Mean Square Error (RMSE), its normalized variant (nRMSE), and the standard deviation of the lateral deviation error (\(\sigma\)) are considered. The lateral deviation error at time step \(t\) is defined as:

\begin{equation*}
e_t = d_t^{\text{pred}} - d_t^{\text{gt}}
\end{equation*}

In which \(d_t^{\text{pred}}\) is the predicted lateral position, and \(d_t^{\text{gt}}\) is the ground-truth lateral position concerning the centerline.

The RMSE is computed as:

\begin{equation*}
\text{RMSE} = \sqrt{ \frac{1}{T} \sum_{t=1}^{T} (d_t^{\text{pred}} - d_t^{\text{gt}})^2 }
\end{equation*}

Where \(T\) is the number of time steps in an episode.

To ensure metric comparability across different contexts, the RMSE is normalized using the lane width \(W_{\text{lane}}\), the mean \(\bar{d}^{\text{gt}}\), or standard deviation \(\sigma_{d^{\text{gt}}}\) of the ground-truth lateral positions:

\begin{equation*}
\text{nRMSE} = \frac{\text{RMSE}}{W_{\text{lane}}}
\end{equation*}

\begin{equation*}
\text{nRMSE} = \frac{\text{RMSE}}{\bar{d}^{\text{gt}}} \quad \text{or} \quad \frac{\text{RMSE}}{\sigma_{d^{\text{gt}}}}
\end{equation*}

The stability of each controller is further evaluated via the standard deviation of lateral errors:

\begin{equation*}
\sigma = \sqrt{ \frac{1}{T} \sum_{t=1}^{T} \left( e_t - \bar{e} \right)^2 }
\end{equation*}

Such that \(\bar{e}\) is the mean of the lateral errors across time. The regression-based validation results, averaged over 50 randomly generated routes, are reported in Table~\ref{tab:lane_rmse}.

\begin{table}[ht]
\centering
\caption{Validation of Lane Tracking Accuracy Using Regression Metrics}
\begin{tabular}{lccc}
\toprule
\textbf{Algorithm} & \textbf{RMSE (m)} & \textbf{nRMSE} & \textbf{Std. Dev. (m)} \\
\midrule
DDPG\cite{perez2022deep}       & 0.42 & 0.120 & 0.22 \\
AR-DDPG\cite{tessler2019action}    & 0.34 & 0.097 & 0.17 \\
AR-RDPG    & 0.27 & 0.077 & 0.14 \\
AR-CADPG   & \textbf{0.23}  & \textbf{0.066} & \textbf{0.11} \\
\bottomrule
\end{tabular}
\label{tab:lane_rmse}
\end{table}

As presented in Table~\ref{tab:lane_rmse}, AR-CADPG achieves the lowest RMSE, nRMSE, and standard deviation among all compared controllers, indicating superior precision and stability in maintaining lane position on snowy and low-friction roads. Incorporating adversarial robustness, temporal context, and attention mechanisms improves control performance and increases system reliability. In contrast, the baseline DDPG shows the highest variation and deviation, indicating reduced stability on slippery surfaces and when lane markings or features are occluded. These results demonstrate that robustness-aware architectural improvements enhance the effectiveness of RL-based lane-keeping systems.

\begin{figure}[htbp]
  \centering
  \includegraphics[width=\linewidth]{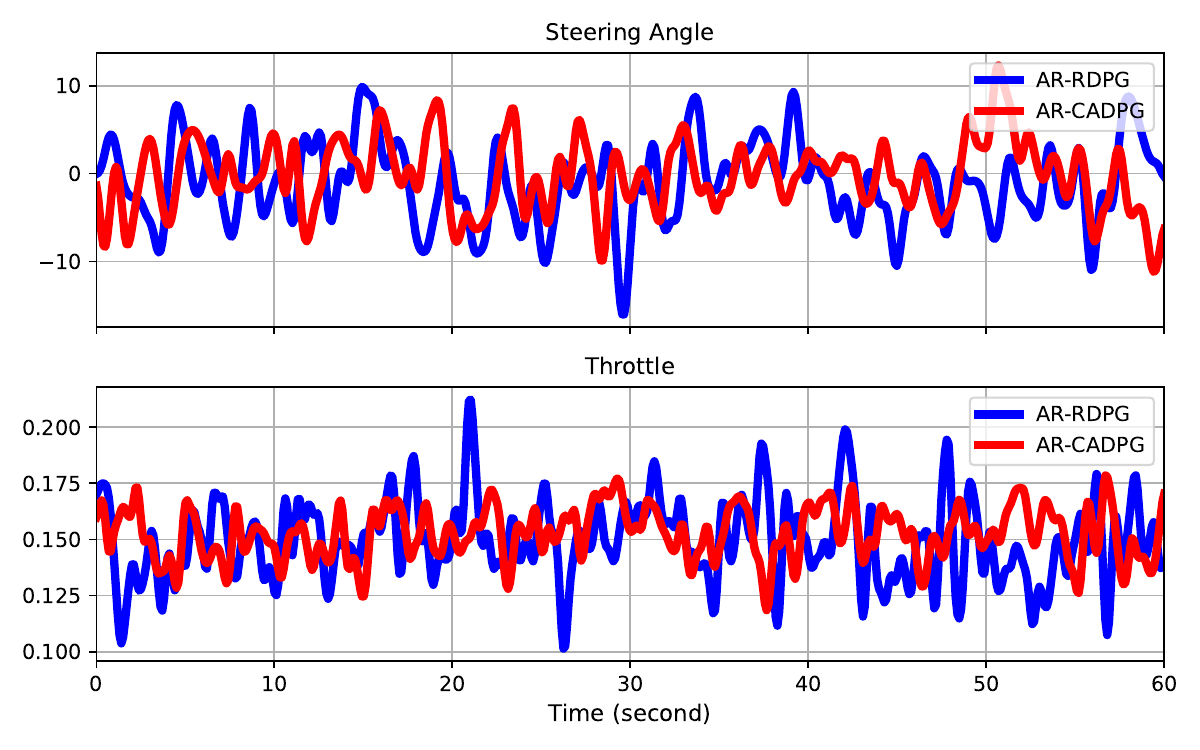}
  \caption{Throttle and steering angle commands produced by AR-RDPG and AR-CADPG controllers on a randomly selected route. Both methods yield smooth and bounded control signals suitable for real-world deployment.}
  \label{fig:command}
\end{figure}

Both the AR-RDPG and AR-CADPG agents record throttle and steering angle commands along a randomly selected test path to evaluate the real-world applicability of the proposed controllers. As illustrated in Fig.~\ref{fig:command}, both controllers generate smooth, continuous, and bounded control signals. No sudden spikes, signal saturation, or high-frequency oscillations are observed, confirming temporal consistency suitable for real-world deployment. The fact that these control outputs remain within operational limits and exhibit stable behavior indicates that the learned policies can be implemented on real-world hardware without additional filtering or post-processing. Furthermore, testing the policies on an unstructured and randomly selected route confirms their robustness and practical applicability.

\section{Testing with a Real-World Robotic Platform}

The proposed lane-keeping controllers are evaluated on a real autonomous vehicle built on the NVIDIA JetRacer platform to assess their real-world performance. This compact robotic car is equipped with a Jetson Nano for processing, a wide-angle HD camera for vision, and an Intel RealSense T265 for visual–inertial odometry. It features a four-core ARM Cortex-A57 CPU and a 128-core Maxwell-based GPU. Together, these components provide sufficient computational power for real-time inference and control.

\begin{figure}[t]
	\centering
	\includegraphics[width=\linewidth]{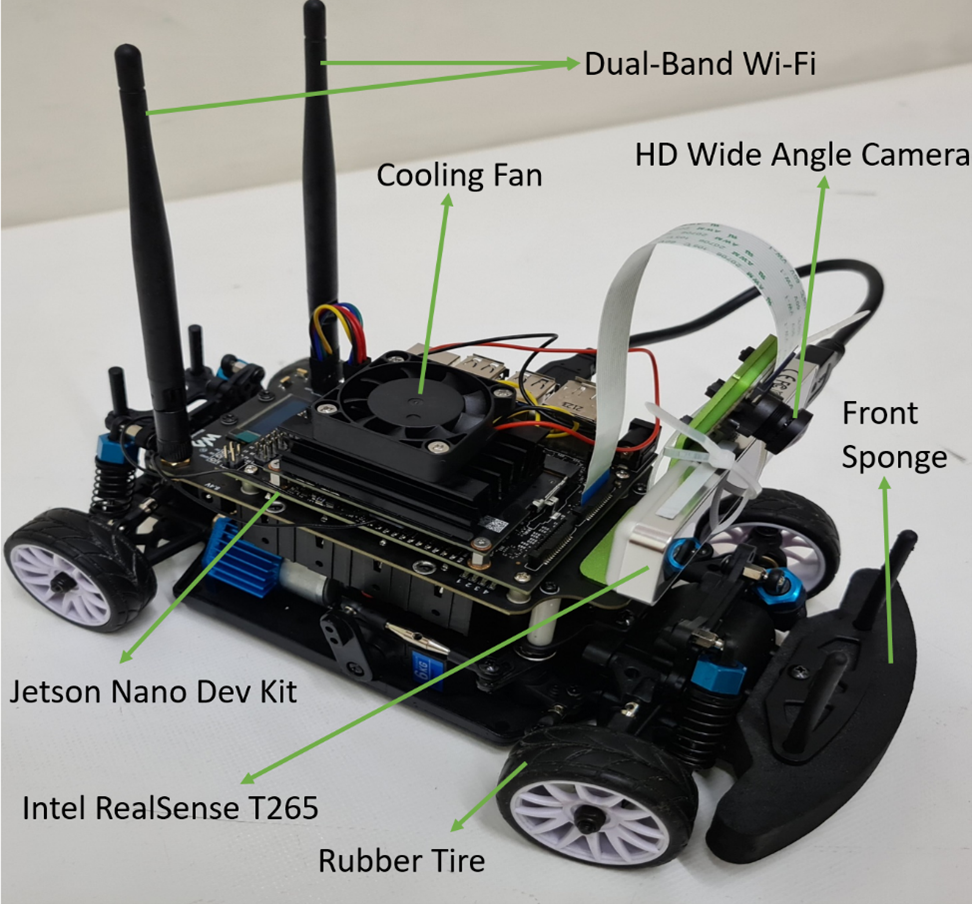}
	\caption{Jetson Nano-based self-driving car used for hardware testing. Main components include the Jetson Nano Dev Kit, RealSense T265, HD wide-angle camera, and wireless modules.}
	\label{fig:robot_overview}
\end{figure}

Fig.~\ref{fig:robot_overview} shows the robotic platform used in the tests. It is equipped with a Jetson Nano Developer Kit, an Intel RealSense T265 tracking camera, a high-definition camera, and wireless communication modules. The JetPack SDK installs a Linux-based operating system on a 128-GB SD card. Python-based libraries such as \texttt{JetCam}, \texttt{torch2trt}, and \texttt{JetRacer} enable real-time perception and control.

The vehicle supports continuous steering and throttle actuation within the range of $[-1, 1]$, where $-1$ and $+1$ correspond to full left/right steering and full braking/acceleration, respectively. These control commands are managed through the \texttt{JetRacer} control stack.

The initial training of the lane-keeping controllers is conducted entirely in simulation. However, simulated environments cannot perfectly replicate how real systems perceive, act, or evolve. Compared with real sensors, cameras in simulation typically differ in resolution, field of view, distortion characteristics, and light sensitivity. The RealSense T265 visual–inertial odometry also exhibits noise patterns and latency that are often absent in most simulation models. Furthermore, actuators in simulation are usually idealized and fail to capture real-world imperfections such as latency, mechanical backlash, or motor variability.

To address these challenges, transfer learning is applied as a key component of the sim-to-real adaptation process. A limited set of real-world driving trajectories is used to fine-tune the controllers initially trained in simulation. This method allows the models to retain the general principles learned during simulation while adapting to the specific characteristics of real sensors and actuators. In practice, this involves modifying the perception and control layers of the DRL policies to handle sensor noise, frame-rate variations, and non-ideal actuation.

\begin{figure*}[!t]
	\centering
	\includegraphics[width=\linewidth]{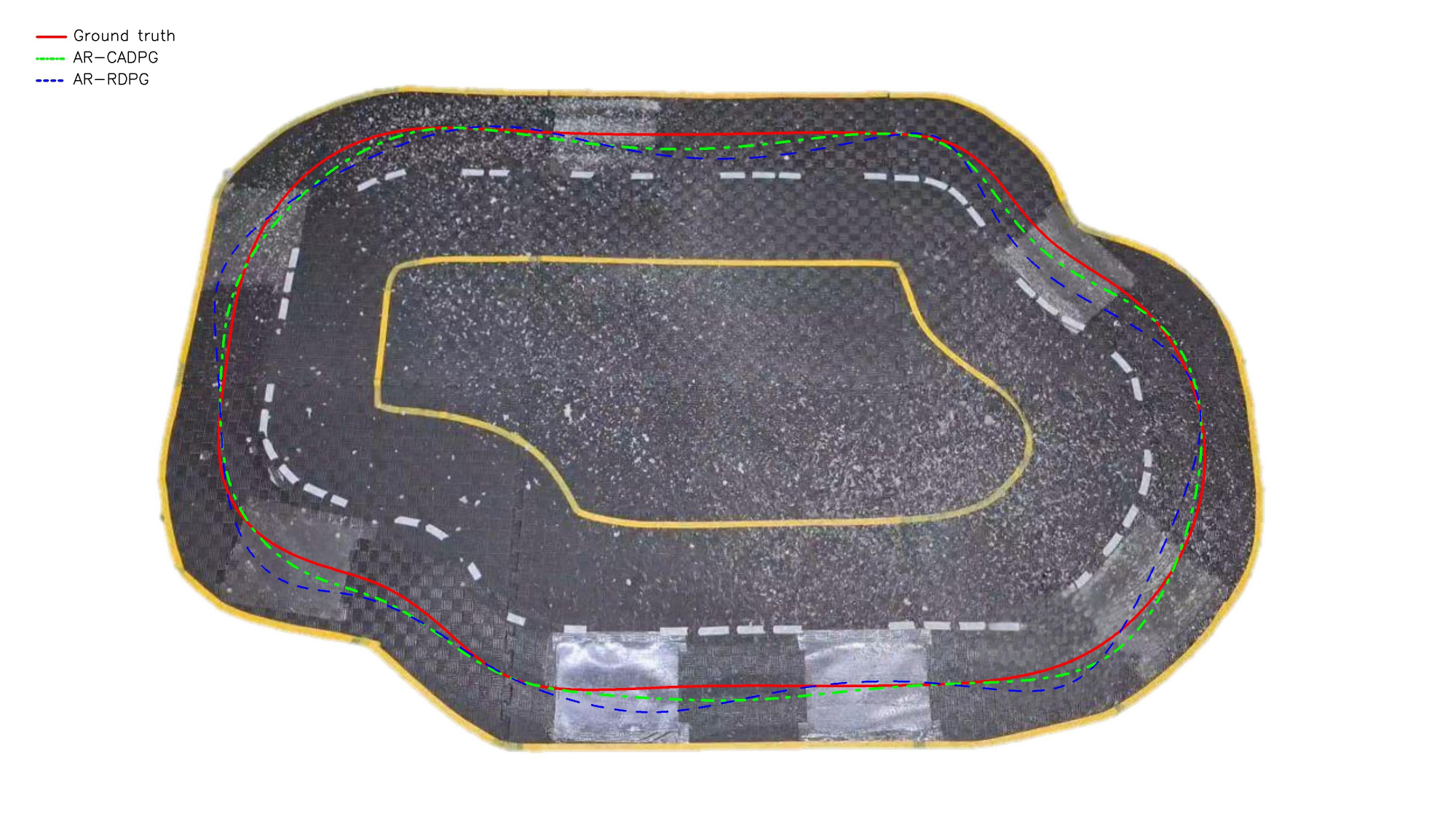}
	\caption{Testing lane-keeping systems in the real world on a real test track.}
	\label{fig:test_track}
\end{figure*}

A dedicated test track is constructed with tatami mats featuring lane markings, as shown in Fig.~\ref{fig:test_track}. The track includes reflective surfaces, occlusion zones, missing lane segments, and “Snow of Joy” particles to degrade perception. Oil patches are added to simulate the low-traction conditions of snowy roads.

These tests demonstrate that the proposed DRL framework is successfully applied to a real-world robotic platform. After fine-tuning, the AR-CADPG controller outperforms the baseline AR-RDPG model, following paths more steadily and recovering from disturbances more effectively. These results align with simulation-based evaluations and confirm that the controller generalizes its learned knowledge to new situations. Transfer learning plays a critical role in bridging the gap between simulation and reality, enabling real-time decision-making while ensuring that the embedded hardware operates within physical constraints.

\section{Conclusion and Future work}

This paper presents a robust DRL framework for LK in snowy and visually degraded driving environments. By integrating advanced perception and control mechanisms, the proposed system addresses key challenges caused by snow-induced uncertainties in both sensor input and vehicle dynamics. A custom reward function guides safe and smooth driving behavior. Two novel DRL algorithms are developed to enhance robustness through temporal memory and adversarial resilience, with AR-CADPG further incorporating attention-based visual processing.

Extensive simulations in the CARLA environment and real-world experiments on a Jetson Nano–based autonomous vehicle demonstrate the effectiveness of the proposed methods. Among the evaluated models, AR-CADPG consistently achieves the highest accuracy, stability, and adaptability under adverse conditions. These results confirm the feasibility of deploying robust, learning-based lane-keeping systems in real-world winter driving scenarios.

Future work explores broader generalization to multi-lane and dynamic traffic environments, integration with additional sensor modalities, and continual learning for long-term adaptation under evolving weather and road conditions.

\bibliographystyle{IEEEtran}
\bibliography{my_bib}

\end{document}